\documentclass{article}
\usepackage{graphicx} % Required for inserting images
\usepackage{tikz}
\usetikzlibrary{positioning}
\usepackage{amsmath,amssymb}
\usepackage{neuralnetwork}
\usepackage{graphicx}
\usepackage{algorithm}
\usepackage[font=tiny]{caption}
\usepackage{mathtools}
\usepackage{hyperref}
\usepackage{blindtext}
\usepackage{multicol}
\usepackage[font=tiny]{caption}
\usepackage{graphicx}
\usepackage{textcomp}
\usepackage{algpseudocode}
\usepackage{subcaption}
\usepackage{amsthm}

\title{Restricted Bayesian Neural Network}
\author{Sourav Ganguly, Saprativa Bhattacharjee}
\date{December 2023}

\begin{document}

\maketitle
\tikzset{%
  every neuron/.style={
    circle,
    draw,
    minimum size=1cm
  },
  neuron missing/.style={
    draw=none, 
    scale=4,
    text height=0.333cm,
    execute at begin node=\color{black}$\vdots$
  },
}
\section*{Abstract}
\textbf{Modern deep learning tools are remarkably effective in addressing intricate problems. However, their operation as black-box models introduces increased uncertainty in predictions. Additionally, they contend with various challenges, including the need for substantial storage space in large networks, issues of overfitting, underfitting, vanishing gradients, and more. This study explores the concept of Bayesian Neural Networks, presenting a novel architecture designed to significantly alleviate the storage space complexity of a network. Furthermore, we introduce an algorithm adept at efficiently handling uncertainties, ensuring robust convergence values without becoming trapped in local optima, particularly when the objective function lacks perfect convexity.}

\textit{Keywords - Bayesian Neural Networks, Cross Entropy Optimization, Deep Learning, Restricted BNN}
\section{Introduction}
The domain of Machine Learning and its applications is expansive. This discipline facilitates the automation and proficient resolution of a myriad of complex problems~\cite{kumar2020machine,dhall2020machine}. At its essence, Machine Learning entails endowing machines with the capability to acquire knowledge and adeptly execute complex tasks~\cite{mohri2018foundations,el2015machine} by learning from examples~\cite{cunningham2008supervised}, finding hidden patterns~\cite{bishop2006pattern}, and trial-and-error~\cite{kaelbling1996reinforcement,arulkumaran2017deep} methodologies.

Machine Learning's impact extends across various sectors, including medicine~\cite{deo2015machine,rajkomar2019machine}, finance~\cite{dixon2020machine,culkin2017machine}, transportation~\cite{tizghadam2019machine,bhavsar2017machine}, and more. A crucial aspect of machine learning involves uncovering underlying patterns within a given dataset. The machine learns these patterns and, by following them, draws valuable conclusions regarding new data points~\cite{bishop2006pattern}.

In many machine learning scenarios, the machine learns a mapping from the input space to the output space. One widely adopted technique for this purpose is constructing a Neural Network (NN)~\cite{abdi1999neural} model with densely connected layers. The NN is essentially a black-box function, that takes inputs from a given dataset. Based on the proximity of the output produced by this black-box function to the true output, it adjusts its weights for continual improvement~\cite{anderson1995introduction}.

While many state-of-the-art algorithms utilize gradient descent to update neural network weights~\cite{du2019gradient}, it may not always provide an accurate solution~\cite{pearlmutter1996investigation}. This is due to the potential presence of local optimum points if the objective function lacks complete convexity~\cite{loh2013regularized}, leading the algorithm to become stuck into a local optimum point. Additionally, advanced neural networks are susceptible to overfitting and underfitting problems~\cite{gavrilov2018preventing}. Various techniques, such as dropout~\cite{srivastava2014dropout,baldi2013understanding} and regularizers~\cite{pereyra2017regularizing}, are employed to mitigate overfitting. In the case of large networks, storing weights layer by layer results in significant storage space requirements.

Despite their potency, deep learning models face the challenge of overfitting, compromising their generalization capabilities. Furthermore, these models often display undue confidence in their predictions, particularly when presenting a confidence interval. This becomes a significant concern in applications where inconspicuous failures could result in substantial consequences, such as in autonomous driving, medical diagnosis, or finance~\cite{izmailov2021bayesian}. Consequently, various approaches have been proposed to tackle this risk. A notable solution is the Bayesian paradigm, which provides a robust framework for analyzing and training neural networks with a heightened awareness of uncertainty. This paradigm extends its support more broadly to the development of learning algorithms that adeptly navigate and comprehend uncertainty~\cite{izmailov2021bayesian}. 

The work presented in \cite{9756596} introduces a Bayesian paradigm, allowing the selection of appropriate priors and sampling of weights from the specified distribution. Subsequently, by adjusting our beliefs gradually based on data samples, the approach is demonstrated. The paper also illustrates the measurement of uncertainty in addressing problems, aiding researchers in connecting their existing knowledge in deep learning with relevant Bayesian methods.

The paper \cite{pmlr-v139-izmailov21a,mackay1995bayesian} provides a noteworthy comparison of Bayesian Neural Networks (BNNs) with other techniques. It demonstrates that BNNs can achieve substantial performance improvements compared to standard training and deep ensembles. The work in \cite{mackay1995bayesian} introduces a novel adaptive model known as the density network. In this neural network, target outputs are provided, but the inputs are unspecified. By placing a probability distribution on the unknown inputs, a latent variable model is defined, capable of discovering the underlying dimensionality of a dataset. The paper also derives and demonstrates a Bayesian learning algorithm for these networks.

In our study, we make the assumption that the weights linked to the incoming connections to the neuron head at a specific layer are sampled from a distribution (specifically, for our work, we have assumed a Normal distribution). Moreover, we refrain from directly adjusting the weights; instead, we employ them to adjust the parameters of the distribution. To train our network without storing the weights, we introduce an algorithm that leverages the Cross Entropy Optimization (CEO)~\cite{botev2013cross} method. We will now provide a summary of the contributions made in this paper.
\begin{itemize}
    \item We introduce an efficient learning method that limits the number of weights (of the Neural network) required to be stored, resulting in significantly reduced space complexity.
\item We present an algorithm that employs the Cross Entropy Optimization (CEO) method to adjust the distribution parameters from which our weights are sampled.
\item We provide a comprehensive comparison with other standard algorithms to highlight the advantages of adopting our proposed method.
\end{itemize}

\section{Background}
This section will explore crucial background details and introduce the necessary notations employed throughout the paper.
\subsection{Neural Networks}
A neural network~\cite{anderson1995introduction} is a black-box function that maps our input feature space to an output space $h: X \to y$ where $X \in \mathbb{R}^{d}$ and $y \in \mathbb{R}^{n}$. Neural networks often consist of multiple layers that receive input from the preceding layer, process the input values (and/or activations through some suitable activation function), and subsequently pass the output to the next layers. 

\begin{figure}
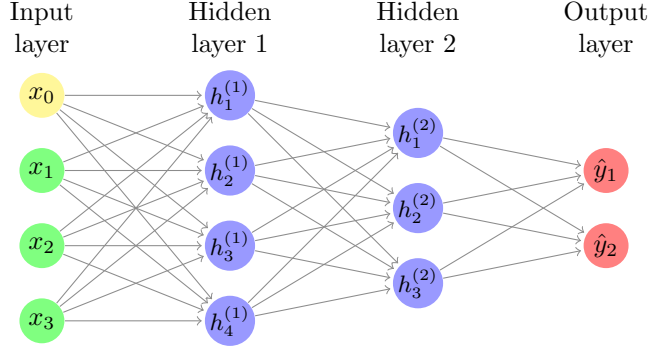

    \centering
   \begin{neuralnetwork}[height=4]
        \newcommand{\x}[2]{$x_#2$}
        \newcommand{\y}[2]{$\hat{y}_#2$}
        \newcommand{\hfirst}[2]{\small $h^{(1)}_#2$}
        \newcommand{\hsecond}[2]{\small $h^{(2)}_#2$}
        \inputlayer[count=3, bias=true, title=Input\\layer, text=\x]
        \hiddenlayer[count=4, bias=false, title=Hidden\\layer 1, text=\hfirst] \linklayers
        \hiddenlayer[count=3, bias=false, title=Hidden\\layer 2, text=\hsecond] \linklayers
        \outputlayer[count=2, title=Output\\layer, text=\y] \linklayers
    \end{neuralnetwork}
    \caption{A simple neural network}
    \label{fig:nn}
\end{figure}

To find the underlying function $h$, we initially consider a surrogate function $h_{s}: X \to \hat{y}$. After obtaining $\hat{y}\in \mathbb{R}^{n}$ for all inputs ($x \in X$), we calculate the closeness between $y$ and $\hat{y}$. Based on the closeness metric, generally referred to as the loss function, we adjust the inner variables, also known as the weights associated with the interconnected layers of the neural network, such that $P(\underset{n \to \infty}{\lim}\hat{y}_{n} \to y_{n}) \to 1$, where $y_{n}$ denotes the outcome after tuning the weights of the neural network for $n-1$ epochs. A simple depiction of a densely connected neural network with one input layer, two hidden layers, and one output layer is illustrated in Figure \ref{fig:nn}.

Let us consider the weights at layer $l$ as $W^{l}$. For our experiments, we mainly focus on Densely connected Neural networks. Let us denote the weight associated with the connection of the $i$th neuron in the $l$th to the $j$th neuron in the $(l+1)$th layer as $W_{i,j}^{l}$ and the output of the weighted sum incoming values after passing through an activation function~\cite{sharma2017activation} to the $j$th neuron be denoted as $o_{j}^{l}$. Further let the number of neurons in the $l$th layer be denoted as $N_{l}$ and $\phi(.)$ be the activation function where, $\phi(.):\mathbb{R} \to \mathbb{R}$.

Let us suppose we obtain an input $x \in \mathbb{R}^{n}$. Using the input vector $x$, we find $h_{s}(x)$. The steps for doing the same are shown below
\begin{equation}
\label{eqn:nn}
\begin{aligned}
    Z^{1}_{j} = \underset{{1\leq i \leq N_{1}}}{\Sigma}W_{i,j}^{1}\cdot x_{i},~\forall~ 1\leq j \leq N_{2},\\
    o^{1}_{j} = \phi(Z^{1}_{j}),~~\forall~ 1\leq j \leq N_{2},\\
    \ldots\\
    Z^{l}_{j} =  \underset{{1\leq i \leq N_{l}}}{\Sigma}W_{i,j}^{l}\cdot o^{l-1}_{i},~\forall~ 1\leq j \leq N_{l},\\
     o^{l}_{j} = \phi(Z^{l}_{j}),~~\forall~ 1\leq j \leq N_{l},
     \end{aligned}
\end{equation}
After passing through all the layers, final outcome is our $\hat{y}$. Thus we can compactly represent $\hat{y}$ as shown in equation \eqref{compact} considering there are $L$ layers in total
\begin{align}
    \hat{y} = \phi(W^{L}\cdot \phi(W^{L-1} \phi(\ldots \phi(W^{1}\cdot x))))
    \label{compact}
\end{align}
The ultimate aim is to get $W^{1},~W^{2} \ldots W^{L}$ so that error between $\hat{y}$ and $y$ is as low as possible ($E: (X,y) \to \mathbb{R}$). Some standard error functions utilized are (i) Mean square error (MSE)~\cite{das2004mean}, (ii) Cross entropy loss function~\cite{ho2019real}, (iii) Mean absolute error (MAE)~\cite{error2016mean} etc. During the training process, we tune the weights such that the true output and the predicted output are as close as possible. This can be done in several ways some of which are mentioned below

\subsubsection{Backpropogation technique}
Most state-of-the-art algorithms are built upon the backpropagation technique~\cite{hecht1992theory}. In this approach, at each epoch, we calculate the error value associated with the current weights. Subsequently, we iteratively adjust the weights in reverse order, moving from the output layer to the input layer. To update the weights, we compute the gradient of the loss or error function with respect to the weights of the $l$th layer, where $1 \leq l \leq L$. Since our goal is to minimize the error, we move in the direction opposite to the gradient and sequentially adjust the weights of each layer. An outline of the backpropagation algorithm is presented in Algorithm \ref{algo:back}.

\begin{algorithm}
\caption{Working of Backpropogation}\label{alg:cap}
\begin{algorithmic}[1]
\State \textbf{Input:} $X,y,N\_epochs,epsilon$
    \State \textbf{Initialize:}$W^{l}~\forall 1\leq l \leq L$ randomly
\For {$i =  1 ~to~ N\_epochs$}
\State Find $\hat{y}$ using equation \eqref{compact}
\State Find error $e=E(y,\hat{y})$
\If $(e>epsilon)$
\For layer = reversed(L,1)
\State Find $\nabla_{W^{layer}} e = \frac{\nabla e}{\nabla_{W^{layer}}}$
\State $W^{layer} = W^{layer} - \nabla_{W^{layer}} e$
\EndFor
\Else 
\State Break inner loop
\EndIf
\EndFor
\State \textbf{Output:} Return $W^{l}~\forall 1\leq l \leq L$
\end{algorithmic}
\label{algo:back}
\end{algorithm}

\subsubsection{Bayesian Neural Network}
%%Study and complete this portion
A Bayesian Neural Network (BNN)~\cite{9756596} is a type of neural network that incorporates Bayesian principles to model uncertainty in its parameters. Unlike traditional neural networks, where weights are treated as fixed values, a BNN represents weights as probability distributions. This allows the model to express uncertainty about the true values of its parameters.

Formally, let's consider a neural network with parameters $\theta$ and input data $X$. In a Bayesian Neural Network, the weights $\theta$ are not fixed values but are treated as random variables. The model defines a prior distribution $P(\theta)$ over the weights, expressing our beliefs about their values before observing any data.

As data is observed, the model updates its beliefs through Bayes' theorem to compute the posterior distribution $P(\theta|X)$, which represents the updated distribution of weights given the observed data. This posterior distribution encapsulates the uncertainty in the model parameters.

During the inference process, predictions are made by integrating over the entire distribution of weights, capturing the uncertainty in the model's predictions. Bayesian Neural Networks are particularly useful in scenarios where uncertainty estimation is crucial, such as in tasks where model predictions impact critical decisions, like medical diagnosis or autonomous systems. The algorithm for BNN is given in algorithm \ref{bnn}

\begin{algorithm}
\caption{Bayesian Neural Network Training}
\label{bnn}
\begin{algorithmic}[1]
\State \textbf{Input:} Training data $X$, labels $Y$
\State \textbf{Initialize:} Prior distribution $P(\theta)$, Neural network architecture
\State Sample initial weights $\theta$ from the prior

\For {each training iteration}
    \State Sample a mini-batch of data $(X_{\text{batch}}, Y_{\text{batch}})$
    \State Compute posterior distribution $P(\theta | X_{\text{batch}}, Y_{\text{batch}})$ using Bayes' theorem
    \State Sample weights $\theta$ from the posterior distribution
    \State Compute neural network predictions for the current mini-batch
    \State Compute the negative log-likelihood loss based on predictions and true labels
    \State Compute the KL divergence between the posterior and prior
    \State Update model parameters using a combination of the negative log-likelihood and KL divergence
\EndFor

\State \textbf{Output:} Trained Bayesian Neural Network

\end{algorithmic}
\end{algorithm}

\subsection{Cross Entropy Optimization technique}
\label{ceo}
As discussed in the preceding section, both Backpropagation and Bayesian Neural Networks often encounter challenges associated with a high number of parameter optimizations, making them unsuitable in certain scenarios due to issues such as vanishing gradients and exploding gradients. It becomes crucial to explore alternative algorithms that do not introduce these problems. Therefore, we turn our attention to a renowned optimization technique known as Cross Entropy Optimization.

The Cross Entropy method (CEM), introduced by Rubinstein~\cite{de2003tutorial}, is a versatile Monte Carlo technique applicable to two types of problems:
\begin{enumerate}
    \item Estimation: It can be used to estimate the value of a function $l=\mathbb{E}\left[H(X)\right]$, where $X$ is a random object taking values in a set $\mathcal{X}$ and H is a function defined on $\mathcal{X}$. A special case of estimation is the estimation of a probability $l=P(S(X)\geq \gamma)$, where S is another function defined on X~\cite{kroese2011cross}.

    \item Optimization: It can be used to optimize (i.e., maximize or minimize) the value of S(x) over all $x \in \mathcal{X}$, where S is an objective function defined on $\mathcal{X}$~\cite{kroese2011cross}.
\end{enumerate}
In the subsequent discussion, we elaborate on the Cross Entropy Method within the optimization context. In a typical optimization problem, our objective is to minimize or maximize a function concerning certain variables, such as $\underset{x}{\max} ~f(x)$. In an unconstrained setting, we typically perform first-order derivatives to identify stationary points and subsequently utilize higher derivatives to determine the stationary point corresponding to the maximum value of $f(x)$.

The Cross Entropy Method (CEM) is an optimization technique that diverges from traditional methods by not relying on the gradient or higher-order derivatives of the objective function. This characteristic makes it particularly suitable for problems where such computations are challenging. Let $f$ be a function $f:\mathbb{R}^d \rightarrow \mathbb{R}$, and let us assume a probability distribution over the input space denoted by $\mathcal{F}$. For simplicity, we assume $\mathcal{F}$ to be a normal distribution, i.e., $\{ x \sim \mathbb{N}(v,\Sigma) \}$. Consider $0 \leq \rho \leq 1$.

We sample $N$ instances from this distribution, denoted as $X_1, X_2, \ldots, X_{N} \sim \mathcal{N}(v,\Sigma)$ where $v \in \mathbb{R}^{d}$, and $\Sigma$ is the covariance matrix. After obtaining all the samples, we evaluate their objective function values $f(X_1), f(X_2), \ldots, f(X_{N})$. Assuming $X_1, X_2, \ldots, X_{N}$ are i.i.d., we sort the samples in descending order based on their corresponding objective function values. Subsequently, we select the first $\lfloor \rho N \rfloor$ samples to update the parameters of our distribution.

Algorithm \ref{cem} outlines the Cross Entropy Method, where we solely update the mean of the parametrized distribution. However, it is also possible to update the covariance matrix.

\begin{algorithm}[H]
\caption{Cross-entropy optimization}\label{cem}
\begin{algorithmic}[1]
\State \textbf{Input:}  $N$, $\rho$, $v_{0} \in \mathbb{R}^d$, $\Sigma$, $T$ 
% \STATE \textbf{Output:} $ind$ - index of change point , $present$ - Boolean indicating presence or absence of change point
%\STATE \textbf{Initialize:}  $j=1$, $v_1 \in \mathcal{V}$, $s_p \in (0,1)$, $\{n_l : 1 \leq l \leq M, n_l > 10 \}$ 
%\STATE D = $\{x_i, ~i \in \{\tau_j+1, \ldots, t\}\}$ (active observation window)
% \STATE $\eta_{\mathcal{A}}= MLE(\mathcal{A}$)
% \STATE Compute the log-likelihood  $LL_0$ of $\mathcal{A}$ under hypothesis $H_0$ from \eqref{LL0}
\For {$j=1$ to $T$} 
\State  Sample $X_1, \ldots, X_{N} \sim \mathbb{N}(v_{j},\Sigma) \}\;\; \textnormal{i.i.d}$
\For {$i=1$ to $N$}
\State Calculate $f(X_{i})$ and store it in a list $L$
\EndFor
%\STATE $H_j$ = POLICY\_UPDATE($\{(V^{d_{0}}_{\pi^{\theta_i}},G_1^{d_{0}}(\pi^{\theta_i})) : 1 \leq i \leq n_j\}$, $\rho_j$, $\alpha_1$)
% \STATE  Sort $\{\theta_i\}_{i=1}^{n_j}$ in ascending order of $G^{d_{0}}$. Let $\xi_j$ be the first $\lfoor \rho_j n_j\rfloor$ elements.
\State Sort L in the descending order
\State Choose the first $\lfloor \rho N \rfloor$ samples and store them in a list $M$
\State $v_{(j+1)} = \alpha \underset{k \in M}{\Sigma}\frac{f(X_{k})X_{k}}{\underset{k \in M}{\Sigma}f(X_{k})} + (1-\alpha) * v_{j}$
\EndFor
\end{algorithmic}
\end{algorithm}
 %The basic idea underlying CEM used is simple. We generate samples of the objective function in each iteration according to some parameterized distribution. A set of elite samples is chosen to correspond to points where the objective function value is high among the samples obtained. Based on these elite samples, the parameters of the distribution from which the points are sampled are updated (see Chapter \ref{Chapter3} for further details).
\section{Problem Formulation}
Having introduced the notations in the previous section, we will now utilize them to articulate the problem we aim to address, in this section. As shown in equation \eqref{eqn:nn}, the output of the weighted sum of $l$th layer goes as input to the $(l+1)$th layer. For our simplicity, we consider the zeroth layer as the input layer and denote it as $o^{0}$ where $o^{0}=x$ ($x\in X$). Thus, compactly, we can write the modified equation as 
\begin{equation}
\label{eqn:twk1}
    \begin{aligned}
        Z^{l}_{j} = \underset{1\leq i \leq N_{l-1}}{\Sigma}W^{l}_{i,j}\cdot o^{l-1}_{i}~, \forall 1 \leq j \leq N_{l},\\
        o^{l}_{j} = \phi(Z^{l}_{j})~,\forall 1 \leq j \leq N_{l},
    \end{aligned}
\end{equation}
Let $w^{l}= \arg \underset{1\leq i \leq N_{l-1}}{\max} |W^{l}_{i,j}|$. Thus, we can tweak equation \eqref{eqn:twk1} to get equation \eqref{eqn:twk2}
\begin{equation}
\label{eqn:twk2}
    \begin{aligned}
         \hat{Z}^{l}_{j} = \underset{1\leq i \leq N_{l-1}}{\Sigma}\frac{W^{l}_{i,j}}{w^{l}}\cdot o^{l-1}_{i}~, \forall 1 \leq j \leq N_{l},\\
        o^{l}_{j} = \phi(\hat{Z}^{l}_{j})~,\forall 1 \leq j \leq N_{l},
    \end{aligned}
\end{equation}
Thus, from equation \eqref{eqn:twk2}, it is safe to consider that all the weights are bounded between $\left[-1,1 \right]$. 
\subsection{Formation of weight matrix}
\begin{figure}[h!]
    \centering
    \begin{subfigure}{0.35\textwidth}
    \begin{tikzpicture}
    \path +(0,2) node[circle,fill=blue!50] (h1) {$o^{1}_{1}$}
    +(0,0) node[circle,fill=red!50] (h2) {$o^{1}_{2}$}
    +(0,-2) node[circle,fill=red!50] (h3) {$o^{1}_{3}$}
    +(-1.5,2) node[circle,fill=green!50] (i1) {$x_{1}$}
    +(-1.5,0) node[circle,fill=green!50] (i2) {$x_{2}$}
    +(-1.5,-2) node[circle,fill=green!50] (i3) {$x_{3}$}
    +(1.3,0) node[circle,fill=yellow!50] (o1) {$o^{2}_{1}$};
    \draw[->] (i1)--(h1)node[pos=.2,below]{$W^{1}_{11}$};
    \draw[->] (i2)--(h1)node[pos=.2,below]{$W^{1}_{21}$};
    \draw[->] (i3)--(h1)node[pos=.2,below]{$W^{1}_{31}$};
    \draw[loosely dotted] (i1)--(h2);
    \draw[loosely dotted] (i2)--(h2);
    \draw[loosely dotted] (i3)--(h2);
    \draw[loosely dotted] (i1)--(h3);
    \draw[loosely dotted] (i2)--(h3);
    \draw[loosely dotted] (i3)--(h3);
    \draw[loosely dotted] (h1)--(o1);
    \draw[loosely dotted] (h2)--(o1);
    \draw[loosely dotted] (h3)--(o1);
    
\end{tikzpicture}
   \caption{Incoming edges to 1st neuron in hidden layer}
    \label{sol:nn1}
        
    \end{subfigure}
\begin{subfigure}{0.3\textwidth}
   \begin{tikzpicture}
    \path +(0,2) node[circle,fill=blue!50] (h1) {$o^{1}_{1}$}
    +(0,0) node[circle,fill=red!50] (h2) {$o^{1}_{2}$}
    +(0,-2) node[circle,fill=red!50] (h3) {$o^{1}_{3}$}
    +(-1.5,2) node[circle,fill=green!50] (i1) {$x_{1}$}
    +(-1.5,0) node[circle,fill=green!50] (i2) {$x_{2}$}
    +(-1.5,-2) node[circle,fill=green!50] (i3) {$x_{3}$}
    +(1.3,0) node[circle,fill=yellow!50] (o1) {$o^{2}_{1}$};
    \draw[->] (i1)--(h2)node[pos=.2,below]{$W^{1}_{12}$};
    \draw[->] (i2)--(h2)node[pos=.2,below]{$W^{1}_{22}$};
    \draw[->] (i3)--(h2)node[pos=.2,below]{$W^{1}_{32}$};
    \draw[loosely dotted] (i1)--(h1);
    \draw[loosely dotted] (i2)--(h1);
    \draw[loosely dotted] (i3)--(h1);
    \draw[loosely dotted] (i1)--(h3);
    \draw[loosely dotted] (i2)--(h3);
    \draw[loosely dotted] (i3)--(h3);
    \draw[loosely dotted] (h1)--(o1);
    \draw[loosely dotted] (h2)--(o1);
    \draw[loosely dotted] (h3)--(o1);
\end{tikzpicture}
    \caption{Incoming edges to 2nd neuron in hidden layer}
    \label{sol:nn2}
\end{subfigure}
\hfill
\begin{subfigure}{0.3\textwidth}
   \begin{tikzpicture}
    \path +(0,2) node[circle,fill=blue!50] (h1) {$o^{1}_{1}$}
    +(0,0) node[circle,fill=red!50] (h2) {$o^{1}_{2}$}
    +(0,-2) node[circle,fill=red!50] (h3) {$o^{1}_{3}$}
    +(-1.5,2) node[circle,fill=green!50] (i1) {$x_{1}$}
    +(-1.5,0) node[circle,fill=green!50] (i2) {$x_{2}$}
    +(-1.5,-2) node[circle,fill=green!50] (i3) {$x_{3}$}
    +(1.3,0) node[circle,fill=yellow!50] (o1) {$o^{2}_{1}$};
    \draw[->] (i1)--(h3)node[pos=.2,below]{$W^{1}_{13}$};
    \draw[->] (i2)--(h3)node[pos=.2,below]{$W^{1}_{23}$};
    \draw[->] (i3)--(h3)node[pos=.2,below]{$W^{1}_{33}$};
    \draw[loosely dotted] (i1)--(h1);
    \draw[loosely dotted] (i2)--(h1);
    \draw[loosely dotted] (i3)--(h1);
    \draw[loosely dotted] (i1)--(h2);
    \draw[loosely dotted] (i2)--(h2);
    \draw[loosely dotted] (i3)--(h2);
    \draw[loosely dotted] (h1)--(o1);
    \draw[loosely dotted] (h2)--(o1);
    \draw[loosely dotted] (h3)--(o1);
\end{tikzpicture}
    \caption{Incoming edges to 3rd neuron in hidden layer}
    \label{sol:nn3}
\end{subfigure}
\hfill
    \caption{Depiction of outgoing edges from input layer}
    \label{fig:nn23}
\end{figure}
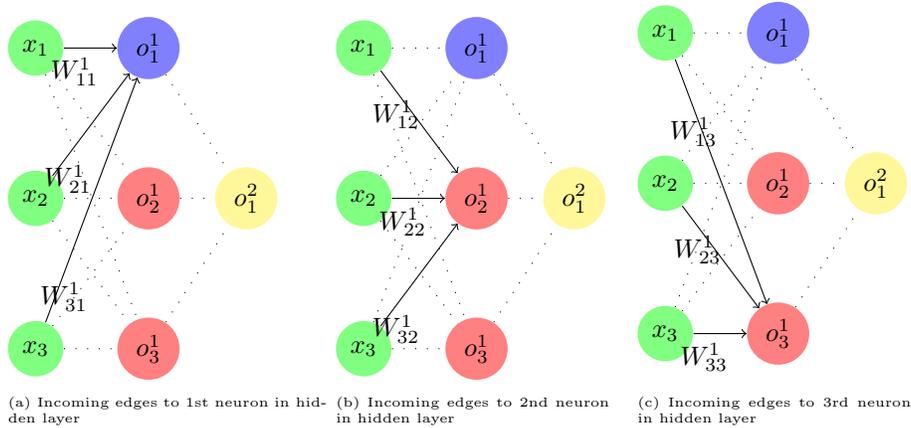

After being able to bind our weights between $\left[-1,1 \right]$, we move on to the formation of the weight matrices between various layers. To do so we consider all the incoming edges to a specific neuron sampled from a distribution (for our experiment we consider Gaussian distribution) as shown in figure \ref{fig:nn23}.

In other words
\begin{equation}
    W^{1} = \begin{bmatrix}
W^1_{11} & W^{1}_{12} & W^{1}_{13}\\
W^1_{21} & W^{1}_{22} & W^{1}_{23}\\
W^1_{31} & W^{1}_{32} & W^{1}_{33}
\end{bmatrix}\\
\text{where}~ W^{1}_{:1} = \begin{bmatrix}
    W^1_{11}\\
    W^1_{21}\\
    W^1_{31}
\end{bmatrix} \sim \mathcal{N}(\mu^{1}_{1},\sigma^{1}_{1})
\end{equation}

Similarly,
\begin{equation}
     W^{1}_{:2} = \begin{bmatrix}
    W^1_{12}\\
    W^1_{22}\\
    W^1_{32}
\end{bmatrix} \sim \mathcal{N}(\mu^{1}_{2},\sigma^{1}_{2})\\
\text{ and}~ W^{1}_{:3} = \begin{bmatrix}
    W^1_{13}\\
    W^1_{23}\\
    W^1_{33}
\end{bmatrix} \sim \mathcal{N}(\mu^{1}_{3},\sigma^{1}_{3})
\end{equation}

\subsection{Optimizing the Gaussian parameters}
After sampling the weights for all the layers, we use them for forward propagation to get $\hat{y}$. After obtaining $\hat{y}$, the error value is computed. Let $f:(X,y) \to \mathbb{R}$ be a composition of two functions $f: E(h_{s}(.)) \to \mathbb{R}$ that is first we find $\hat{y}$ by passing the input through the black-box Neural network and then we calculate the error value using the error function or cost function for all the data entries. We then repeat these steps for $n$ number of samples and for each weight samples, the total error is stored in a list. Following this an elite group of samples having least total error is selected (for our experiments we have chosen 10\% of total samples as elite samples) and use these samples to update our mean and sigma values. In this way we tune the weights using Cross-Entropy optimization. The complete algorithm is given in Algorithm \ref{rbnn}

\begin{algorithm}[H]
\caption{Restricted BNN}\label{rbnn}
\begin{algorithmic}[1]
\State \textbf{Input:}  $n_{j}$ which a list n values for j iterations, $\tau$ the number of iterations we sample weights and perform updation $\mu\in \mathbb{R}^{(L,2)}$ the mean value,$\sigma$, the fixed standard deviation and $L$ denoting number of layers, $X$ denoting the input data, $y$ denoting given output label $\rho$ the inter-quartile value

\For {$j=1$ to $\tau$} 
\For {$l=1$ to $L-1$}
\State  $W_{:k} \sim \mathcal{N}(\mu^{l}_{(k)},\sigma,N_{l})~\forall 1\leq k \leq N_{(l+1)}$ (So it generate $N_{l}$ samples from a Normal distribution with $\mu^{l}_{(k)}~,~\sigma$ parameters)
\State $W^{l}$ in a dictionary $W$
\EndFor
\State $error=0$
\For {$i=1$ to $N$}
\State Calculate $err = (y_{i} - forward(X_{i},W))$ and store it in a list $L$ (See algorithm \ref{forward} for forward subroutine)
\State $error = error + err$
\EndFor
\State Store error in list $\mathbb{M}$
%\STATE $H_j$ = POLICY\_UPDATE($\{(V^{d_{0}}_{\pi^{\theta_i}},G_1^{d_{0}}(\pi^{\theta_i})) : 1 \leq i \leq n_j\}$, $\rho_j$, $\alpha_1$)
% \STATE  Sort $\{\theta_i\}_{i=1}^{n_j}$ in ascending order of $G^{d_{0}}$. Let $\xi_j$ be the first $\lfoor \rho_j n_j\rfloor$ elements.
\State Sort $\mathbb{M}$ in the ascending order
\State Choose the first $\lfloor \rho n_{j} \rfloor$ samples and store them in a list $M$
\State $v_{(j+1)} = \alpha \underset{k \in M}{\Sigma}\frac{f(X_{k})X_{k}}{\underset{k \in M}{\Sigma}f(X_{k})} + (1-\alpha) * v_{j}$
\EndFor
\end{algorithmic}
\end{algorithm}

\begin{algorithm}[H]
    \caption{forward(x,W)}\label{forward}
\begin{algorithmic}[1]
    \State $o^{0}=x$
    \For {$i=1$ to $L$}
    \State $o^{i} = \Phi( W^{T}\cdot o^{l-1})$ ($\Phi(.)$ takes a vector and returns a vector of $\phi(w^{T} \cdot o^{l-1}) $where $w$ are columns of $W$)
    \EndFor
    \State \textbf{Return:} $o^{L}$
\end{algorithmic}
\end{algorithm}

\section{Results}
We evaluated the performance of our algorithm on various benchmarks. Additionally, we conducted comparisons with a similarly structured feed-forward neural network and a conventional Bayesian neural network.
\subsection{Benchmarks}
The benchmark datasets employed to assess the models are all standard classification-based datasets. While this approach can be extended to regression-type models and time-series models, our study confines itself to performing classification tasks using this model.
\subsubsection{Pulsar-star dataset}
The Pulsar Star dataset is a collection of data related to the identification of pulsars, which are a specific type of celestial object. Pulsars are highly magnetized, rotating neutron stars that emit beams of electromagnetic radiation out of their magnetic poles. These beams of radiation are observed as periodic pulses of light when they are aligned with the Earth.

The Pulsar Star dataset is often used in machine learning and data mining research for binary classification tasks. The objective is to distinguish between pulsar stars and other sources of electromagnetic signals. The dataset typically contains features derived from pulsar candidates and non-pulsar candidates, with the target variable indicating whether a candidate is a pulsar or not.

Key attributes in the dataset may include various statistical measures and other characteristics of the candidate signals. Researchers and data scientists use this dataset to build and evaluate classification models, such as binary classifiers, to automatically identify pulsar stars based on the provided features. 

\begin{figure}[h]
     \centering
     \begin{subfigure}[b]{0.45\textwidth}
         \centering
         \includegraphics[width=\textwidth]{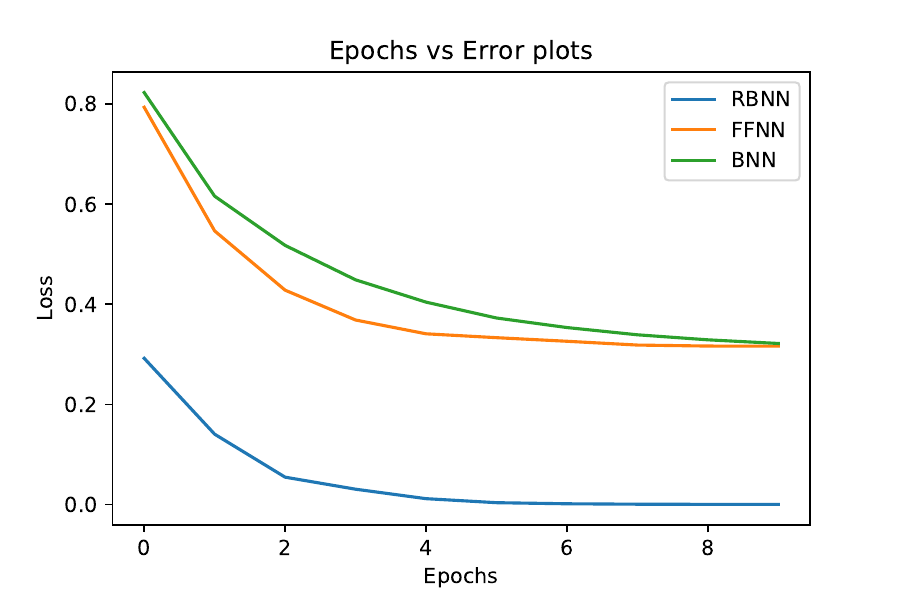}
         \caption{Comparison plot of errors between Restricted BNN, FFNN and BNN}
         \label{fig:pul_comp}
     \end{subfigure}
     \hfill
          \begin{subfigure}[b]{0.45\textwidth}
         \centering
         \includegraphics[width=\textwidth]{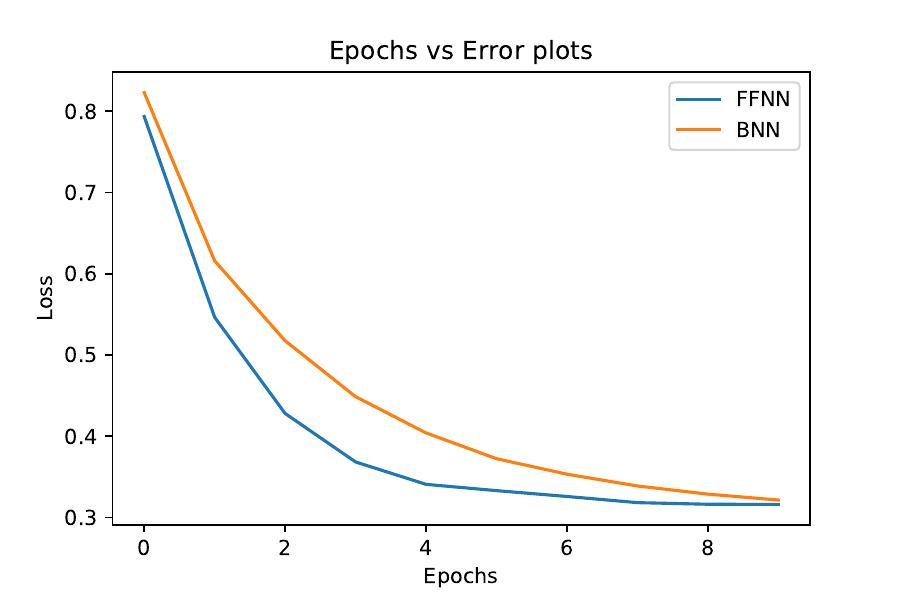}
         \caption{FFNN vs BNN loss plot}
         \label{fig:puls}
     \end{subfigure}
     \hfill
        \caption{Error plots on Pulsar star dataset}
        \label{fig:pulsar_plot}
\end{figure}

The results obtained on testing our algorithm on this dataset and on comparing with FFNN and BNN are shown in figures \ref{fig:pul_comp} and \ref{fig:puls}

The final accuracy obtained on testing on a subset (100 data rows) of the dataset on all the three algorithms is as depicted in table \ref{tab:comp_pulsar}

\begin{table}[h!]
    \centering
    \begin{tabular}{|c|c|c|}
    \hline
       \textbf{Sl. no.}  &\textbf{Neural Network Model}&\textbf{Accuracy obtained}  \\
       \hline
        1 & FFNN & 0.907625\\
        \hline
        2 & BNN & 0.90124\\
        \hline
        3 & RBNN & 0.92865\\
        \hline
    \end{tabular}
    \caption{Comparison of accuracies on same subset of dataset upon all three algorithms}
    \label{tab:comp_pulsar}
\end{table}
\subsubsection{Iris-dataset}
\begin{figure}
     \centering
     \begin{subfigure}[b]{0.45\textwidth}
         \centering
         \includegraphics[width=\textwidth]{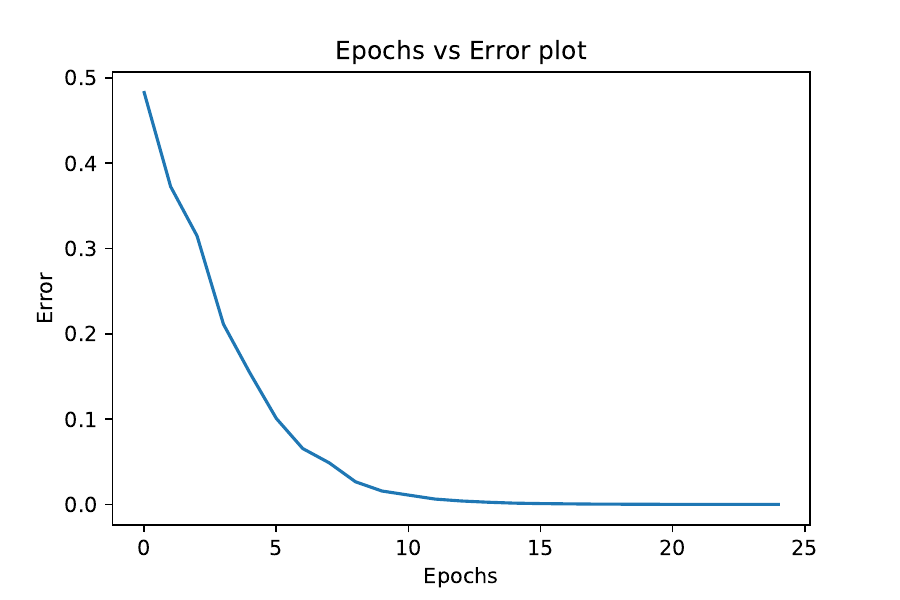}
         \caption{Total error plot during training}
         \label{fig:iris}
     \end{subfigure}
     \hfill
     \begin{subfigure}[b]{0.45\textwidth}
         \centering
         \includegraphics[width=\textwidth]{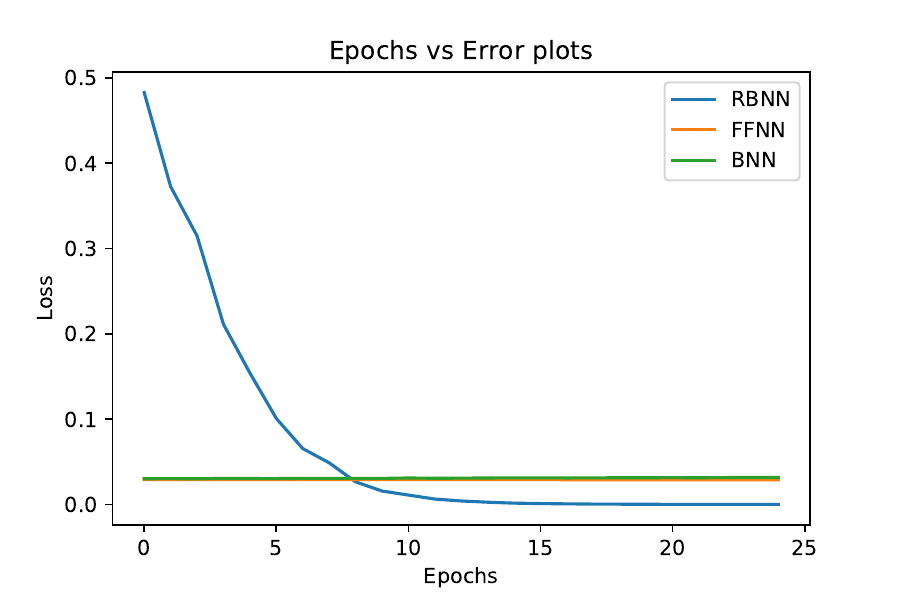}
         \caption{Comparison plot of errors between Restricted BNN, FFNN and BNN}
         \label{fig:iris_comp}
     \end{subfigure}
     \hfill
     \begin{subfigure}[b]{0.45\textwidth}
         \centering
         \includegraphics[width=\textwidth]{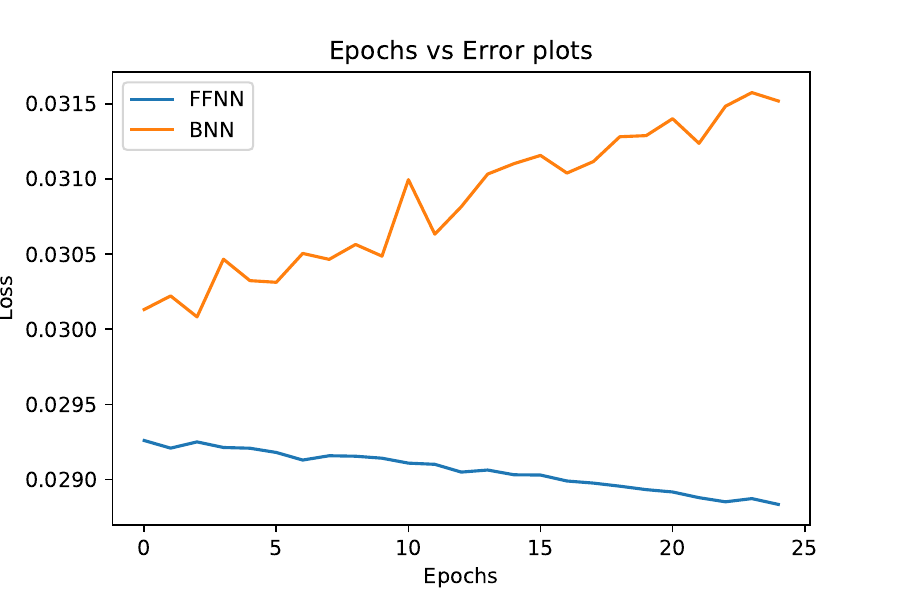}
         \caption{Comparison plot of errors between FFNN and BNN}
         \label{fig:ffnn vs bnn}
     \end{subfigure}
        \caption{Error plots on IRIS dataset}
        \label{fig:iris_plot}
\end{figure}

The Iris dataset, curated by British biologist and statistician Ronald A. Fisher in 1936, holds significance in the realms of both machine learning and statistics. Frequently employed as a benchmark in machine learning applications, this dataset serves as a standard for tasks involving classification and clustering. Comprising measurements of four features (sepal length, sepal width, petal length, and petal width) for a total of 150 iris flowers, the dataset encapsulates three distinct species: setosa, versicolor, and virginica. Each species is represented by 50 samples. Primarily utilized to hone and showcase various machine learning algorithms, the Iris dataset is especially prevalent in the domain of supervised learning. In this context, the dataset is often employed to develop models that classify iris flowers into their respective species based on the recorded features. The training error plot has been shown in figure \ref{fig:iris}. The complete comparison with FFNN and BNN is as shown in figure \ref{fig:iris_comp} and \ref{fig:ffnn vs bnn}

\subsection{Discussion}
We train our model over two standard datasets namely IRIS dataset and the Pulsar-star dataset. Moreover, we compared the loss plots of our model with the standard feed-forward neural networks and bayesian neural networks as shown in figures \ref{fig:pul_comp} and \ref{fig:iris_comp}. Pulsar dataset is a very large dataset consisting of nearly twenty-thousand data samples. However, the dataset is filled with missing data columns. For our experiments we removed all the missing columns and thus, reduced the data to 12528 data samples. We also dropped the text data present in the columns and the reduced pulsar data-set was used for training in all the three models. The plots essentially is a comparison of how fast the total loss goes down to a minimum value. The observations made from this experiment are as follows
\begin{enumerate}
    \item RBNN model has proved to be much more efficient in terms of storage space and accuracy as compared to FFNN and BNN. Within 10 epochs the loss has moved down to almost zero in RBNN. 
    \item The accuracy obtained while using RBNN on a subset of the dataset is nearly 93\%  whereas it is nearly 91\% and 90\% while using FFNN and BNN respectively\footnote{The complete code along with the accuracy and plots is given in \url{https://github.com/Sourav1429/Restricted_BNN.git}}.
    \item The storage space required for this model is also lesser as compared to FFNN and BNN. For RBNN only the means required to be stored, for FFNN the complete weights are required to be stored whereas in BNN the means and the standard deviation associated with each connection was required to be stored. (structure given in table \ref{comp_table1})
\end{enumerate}

After getting some good results on the Pulsar dataset, a genuine doubt was how good our model can perform on multi-class classification problem. This is because in many real-life problems there are multiple classes and it is crucial to find out the performance of RBNN on multi-class classification problems. In order to do this, we take a standard dataset called the IRIS dataset. Like pulsar here again we compare it with FFNN and BNN. The good thing was in this problem there are no missing values so not much pre-processing was necessary. The labels however, were converted into one-hot vectors for efficient results and then passed onto all our models. The observations made from this experiment are as follows
\begin{enumerate}
    \item RBNN model initially had a very high error as compared to BNN and FFNN. This is because initially the weights are sampled randomly from distributions whose mean was far away from the true mean values. However, later upon training for few epochs (precisely 10 epochs), our model proved much more efficient as compared to FFNN and BNN. Thus, showing that this black-box function can give far better results than FFNN and BNN even on multi-class classification problems.
    \item The accuracy obtained after testing on a subset of the dataset (10 samples) it was observed that RBNN achieve an accuracy of 99.3\% whereas FFNN and BNN achieved an accuracy of 95.8\% and 92.6\% respectively. 
    \item The storage space required for this model is also lesser as compared to FFNN and BNN. For RBNN only the means required to be stored, for FFNN the complete weights are required to be stored whereas in BNN the means and the standard deviation associated with each connection was required to be stored. (structure given in table \ref{comp_table2})
\end{enumerate}
A detailed comparison and analysis of the results for the three models\footnote{The machine these models were tested on was having Windows 11 operating system, Intel i7 processor, 128 GB SSD, no GPU and coded using jupyter notebook} are given in table \ref{comp_table1} and table \ref{comp_table2}

\begin{table}[h!]
    \centering
    \begin{tabular}{|c|c|c|c|c|}
    \hline
         \textbf{Model name} & \textbf{Parameters stored} & \textbf{Accuracy (in \%)} & \textbf{Epochs}& \textbf{Time taken}  \\
         \hline
         RBNN&6 parameters(\{2,2,2\})&92.865&10&15 minutes\\
         \hline
         FFNN&12 parameters(\{4,4,4\})&90.7625&10&2 minutes\\
         \hline
         BNN&24 parameters (\{8,8,8\})&90.124&10&3 minutes\\
         \hline
    \end{tabular}
    \caption{Comparison of various observations between RBNN, FFNN and BNN on Pulsar dataset}
    \label{comp_table1}
\end{table}

\begin{table}[h!]
    \centering
    \begin{tabular}{|c|c|c|c|c|}
    \hline
         \textbf{Model name} & \textbf{Parameters stored} & \textbf{Accuracy (in \%)} & \textbf{Epochs}& \textbf{Time taken}  \\
         \hline
         RBNN&8 parameters(\{4,2,2\})&99.3&25&3 minutes\\
         \hline
         FFNN&16 parameters(\{8,4,4\})&95.8&25&38 seconds\\
         \hline
         BNN&32 parameters (\{16,8,8\})&92.6&25&42 seconds\\
         \hline
    \end{tabular}
    \caption{Comparison of various observations between RBNN, FFNN and BNN on IRIS dataset}
    \label{comp_table2}
\end{table}

\subsection{Advantages and Limitations of this model}
The proposed model has demonstrated comparable performance to both the FFNN and BNN. One key advantage lies in the significant reduction of space complexity, as we don't store the weights in the model. At each step, we sample weights from a distribution characterized by mean and standard deviation, eliminating the need for weight storage. This becomes particularly beneficial in large models, where storage requirements can be substantial. In our approach, this demand for extensive storage space is circumvented.

The cross-entropy optimization technique, being a zero-order optimization method, eliminates the need for computing gradients and higher-order derivatives. Additionally, it guarantees convergence to a global minimum. The absence of gradients in this model addresses issues related to vanishing and exploding gradients during weight updates. Furthermore, the model's inherent random sampling from a distribution removes the necessity for dropout, as it avoids introducing bias towards any neuron head.

Despite these advantages, the model does have notable limitations. Convergence to the optimal solution may take a considerable amount of time, especially when the sampled values are far from the correct weights. The convergence time is dependent on the size of the model, and constructing larger models may result in longer durations to reach optimal weight values.

\section{Conclusions and Future work}
In this study, we introduced an efficient Restricted Bayesian Neural Network model trained using the Cross-Entropy method. Our proposed algorithm and model significantly reduce storage space complexity for large models and effectively address issues related to overfitting and underfitting. Notably, our approach eliminates the need for calculating higher-order gradient values, simplifying the solution process. The experimental results highlight the efficiency and practical applicability of our model.

Looking ahead, exploring regression models and time-series models represents an intriguing path for future research. Extending our algorithms to accommodate larger models, and enhancing efficiency in terms of space considerations, stands out as another compelling direction. Additionally, expediting the algorithm to achieve faster convergence emerges as a vital research focus, enhancing the robustness and applicability of the algorithm, especially in larger models with numerous layers and neurons.

\section*{Acknowledgements}
I extend my sincere gratitude to Mr. Abhinaba Ghosh, Ph.D. student at IIT Bombay, for their invaluable support throughout this project.

\bibliographystyle{IEEEtran} % Use the "custom" BibTeX style for formatting the Bibliography

%\bibliography{biblography} % The references (bibliography) information are stored in the file named "Bibliography.bib"
\end{document}